\documentclass{dynafront2026} 

\title[Slot Attention via Particle Dynamics]{Understanding Clustering in Slot Attention via Particle Dynamics}

\usepackage{booktabs}
\usepackage{cleveref}
\usepackage{subcaption}

\newcommand{\R}{\mathbb{R}}

\optauthor{%
\Name{Vasudev Joy*} \Email{vasudev.joy22@imperial.ac.uk}\\
\Name{Rajat R Rasal*} \Email{rajat.rasal17@imperial.ac.uk}\\
\Name{Avinash Kori} \Email{a.kori21@imperial.ac.uk}\\
\Name{Anthea Monod} \Email{a.monod@imperial.ac.uk}\\
\Name{Ben Glocker} \Email{b.glocker@imperial.ac.uk}\\
\addr Imperial College London}

\begin{document}

\maketitle

\vspace{-3em}
\begin{abstract}
Studying attention through the lens of interacting particle dynamics has shown how token clustering can emerge from the underlying dynamics.
We extend this perspective to \emph{slot attention}, a method for object-centric image segmentation and representation learning in which learned components obscure how much of the clustering behaviour is intrinsic to the attention dynamics. 
We therefore introduce \emph{simplified slot attention} (SSA), a parameter-free variant whose dynamics are connected to soft $k$-means clustering and which provides a straightforward mechanistic explanation for the emergence of object-centric representations.
On the Pascal VOC dataset, SSA achieves performance comparable to that of slot attention, demonstrating that competitive object-centric segmentation can be achieved without learned neural-network components.

\end{abstract}

\section{Introduction}

A growing body of theoretical work studies attention as an interacting particle system, where tokens are particles whose dynamics are governed by repeated attention updates \citep{geshkovski_emergence_2024, karagodin_clustering_2024, chen_quantitative_2025, rigollet_mean-field_2026}.
These dynamics have been shown to produce emergent token clustering.
In particular, \citet{geshkovski_mathematical_2025} set the query, key, and value projections to the identity, which enables them to characterise the resulting clustering dynamics and analyse the associated energy functions.

We apply this perspective to \emph{slot attention} \citep{locatello_object-centric_2020}, a method for object-centric representation learning that groups image features into a fixed set of latent representations, called \emph{slots}, intended to capture objects in a scene.
Its competitive assignment mechanism produces segmentation masks, a behaviour \citet{locatello_object-centric_2020} informally relate to soft clustering.
However, the model's learned components obscure how much of this clustering is intrinsic to the attention dynamics themselves.
Here, we ask whether this competition alone is sufficient for object-centric segmentation.

Our main contributions are as follows:
(1) we formulate slot attention as an interacting particle system and derive simplified slot attention (SSA), a parameter-free model that isolates its competitive assignment dynamics;
(2) we relate SSA to soft $k$-means, soft spherical $k$-means, and USSA, an unnormalised energy-based surrogate for SSA which we derive; 
(3) on Pascal VOC, SSA achieves segmentation and reconstruction performance close to slot attention, suggesting that much of slot attention's clustering can be explained by its dynamics rather than its learned components.

\section{Preliminaries}

Let \(X=[x_1,\ldots,x_N]^\top\in\R^{N\times d}\) denote the input features and \(S^t=[s_1^t,\ldots,s_M^t]^\top\in\R^{M\times d_{\mathrm{slots}}}\) the slots at iteration \(t\). 
Throughout, \(Q,K,V\) denote the query, key, and value projection matrices, while \(w_{ij}^t\) denotes the attention or assignment weight from representation \(i\) and input \(j\).

\subsection{Slot Attention} 
\label{sec:slot_attn}
Slot attention \citep{locatello_object-centric_2020} groups encoder features into \(M\) latent representations, called \emph{slots}.
In our setup, a frozen DINO encoder \citep{caron_emerging_2021} maps an image to patch features \(X\), which slot attention groups into slots \(S^t\).
Queries are obtained from the slots, while keys and values are obtained from the fixed encoder features.
We use \(Q\in\R^{p\times d_{\mathrm{slots}}}, K\in\R^{p\times d}, V\in\R^{d_{\mathrm{slots}}\times d}\), so slot attention is written as
\begin{equation}
    \label{slot-attn}
    \operatorname{SlotAttention}(S^t,X)
    =
    \operatorname{WeightedMean}\left[
        \operatorname{Softmax}_{c}\left(
            \frac{(S^tQ^\top)(XK^\top)^\top}{\sqrt{p}}
        \right)
    \right]XV^\top ,
\end{equation}
where \(\operatorname{Softmax}_{c}\) is applied column-wise over slots and \(\operatorname{WeightedMean}\) normalises each row by its total assignment weight.
Thus, the \(M\) slots compete to explain each input feature.
Input features are layer normalised before iteration, \(\hat X=\operatorname{LN}(X)\), while slots are normalised at each iteration, \(\hat S^t=\operatorname{LN}(S^t)\).
They are updated by a gated recurrent unit (GRU) \citep{cho_learning_2014} followed by a residual MLP,
\begin{align}
    S^{t+1}
    &=
    \tilde S^{t+1}
    +
    \operatorname{MLP}\left(
        \operatorname{LN}(\tilde S^{t+1})
    \right),
    \qquad
    \label{slot-attn-update}
    \tilde S^{t+1}
    =
    \operatorname{GRU}\left(
        S^t,\;
        \operatorname{SlotAttention}(\hat S^t,\hat X)
    \right).
\end{align}
Following \citet{seitzer_bridging_2023}, a decoder maps the slots back to reconstructions of the DINO patch features and corresponding alpha masks, which weight a slot's contribution to the reconstruction.
Segmentation masks are obtained from either the assignments in \cref{slot-attn} or the decoder alpha masks.

\subsection{Particle Dynamics}
\label{sec:particle_dynamics}

\paragraph{Cross-Attention Dynamics.}
Following \citet{geshkovski_mathematical_2025}, repeated cross-attention layers can be indexed by time \(t\) and written as the interacting particle system
\begin{equation}
    \label{eq:cross_attn}
        s_i^{t+1}
        =
        s_i^t + V \sum_{j=1}^N w_{ij}^t x_j,
        \qquad
        w_{ij}^t
        =
        \frac{
            e^{\beta\langle Qs_i^t, Kx_j\rangle}
        }{
            \sum_{l=1}^N e^{\beta\langle Qs_i^t, Kx_l\rangle}
        }.
\end{equation}
where \(\beta>0\) controls the sharpness of the attention weights.
Akin to slot attention, all the slots here evolve with respect to the fixed input features.
The weights are normalised over the input index \(j\), \(\sum_j w_{ij}^t = 1\), so the corresponding attention matrix, \((W^t)_{ij}=w^t_{ij}\), is row-stochastic.

\paragraph{Soft \(k\)-Means (SKM) Dynamics.}
\citet{bauckhage_lecture_2015} updates \(M\) cluster centroids using
\begin{equation}
    \label{eq:SKM}
    s^{t+1}_i
    =
    \frac{\sum_{j=1}^N w^t_{ij}x_j}
         {\sum_{j=1}^N w^t_{ij}},
    \qquad
    w^t_{ij}
    =
    \frac{
        e^{-\beta\|s_i^t-x_j\|^2}
    }{
        \sum_{m=1}^M e^{-\beta\|s_m^t-x_j\|^2}
    }.
\end{equation}
where \(\beta\rightarrow\infty\) recovers the standard $k$-means. Constraining data and centroids to a sphere gives us soft spherical \(k\)-means (SSKM)
\begin{equation}
    \label{SSKM}
    s^{t+1}_i
    =
    \frac{\sum_{j=1}^N w_{ij}^t x_j}
         {\left\|\sum_{j=1}^N w_{ij}^t x_j\right\|},
    \qquad
    w_{ij}^t
    =
    \frac{
        e^{2\beta\langle s_i^t,x_j\rangle}
    }{
        \sum_{m=1}^M e^{2\beta\langle s_m^t,x_j\rangle}
    }.
\end{equation}
Unlike cross-attention, SKM and SSKM normalise over the cluster index \(i\), \(\sum_{i=1}^M w_{ij}^t=1\), so their assignment matrices are column-stochastic.
Hence, centroids compete for each input.

\section{Simplified Slot Attention}
\label{sec3}

\subsection{Dynamical Model}
\label{simplification}

We now formulate slot attention as an interacting particle system, and derive a parameter-free dynamical model that isolates its competitive assignment mechanism.
Akin to \cref{eq:cross_attn}, slot attention can be written as, 
\begin{gather}
    \label{SA-1}
    s^{t+1}_i = \tilde{s}^{t + 1}_i + \operatorname{MLP}\left(\operatorname{LN}(\tilde{s}^{t+1}_i)\right), \qquad
    \tilde{s}^{t+1}_i = \operatorname{GRU}\left(s^t_i,\; V \frac{\sum_{j = 1}^N w_{ij}^tx_j}{\sum_{j = 1}^N w_{ij}^t}\right), \quad \\
    \text{where} \quad 
    w_{ij}^t= \frac{e^{\beta\langle Qs_i^t, Kx_j\rangle}}{\sum_{m = 1}^M e^{\beta\langle Qs^t_m, Kx_j \rangle}}.
\end{gather}
Two reductions bring this into the form of a particle system.
First, following \citet{geshkovski_mathematical_2025} and \citet{locatello_object-centric_2020}, we assume \(p = d = d_\text{slots}\) and set \(Q=K=V=I\), so that slots and inputs interact directly through their inner products.
Second, we take the weighted-mean aggregation as the update itself, in place of the LayerNorm, GRU, and residual MLP, leading to
\begin{equation}
    \label{eq:ssa}
    s^{t+1}_i = \frac{\sum_{j = 1}^N w_{ij}^tx_j}{\sum_{j = 1}^N w_{ij}^t} ,     \qquad w_{ij}^t= \frac{e^{\beta\langle s_i^t, x_j\rangle}}{\sum_{m = 1}^M e^{\beta\langle s^t_m, x_j \rangle}},
\end{equation}
which we refer to as simplified slot attention (SSA).
SSA retains slot attention's column-stochastic assignments, \(\sum_{i} w^t_{ij} = 1\), so the \(M\) slots compete to explain each input \(x_j\).
Being parameter-free, it lets us ask how much of slot attention's clustering the competitive mechanism already accounts for.
We interpret the omitted components in \Cref{excluded-components}.

\subsection{Relationship to Soft Clustering Methods}
\label{relationship-to-kmeans}

We now characterise the relationship between the competitive assignment dynamics of SSA and the clustering methods introduced in \Cref{sec:particle_dynamics}.
Both SSA and SKM use column-stochastic weights and update each slot or centroid using a weighted mean of the input points.
They therefore share the same competitive assignment structure, but differ in their similarity functions, 
\begin{equation}
    \text{SKM:} \quad w_{ij}^t \propto e^{-\beta\|s_i-x_j\|^2}
    \qquad 
    \text{SSA:} \quad w_{ij}^t \propto e^{\beta' \langle s_i,x_j\rangle}.
\end{equation}
When slots and inputs are restricted to the unit sphere, as in SSKM (\cref{SSKM}), these assignment rules are equivalent for \( \beta' = 2 \beta \) \citep{krimmel_attention_2024}. 
They do, however, differ in their updates, 
\begin{equation}
    \text{SSKM:} \quad s_i^{t+1} = \mathbf{C} \bigg\lVert \sum_j w_{ij}^t x_j \bigg\rVert^{-1}
    \qquad
    \text{SSA:} \quad s_i^{t+1} = \mathbf{C} \bigg( {\sum_j w_{ij}^t} \bigg) ^{-1},
\end{equation}
with $\mathbf{C} =\sum_j w_{ij}^t x_j$.
SSKM projects each updated centroid onto a unit sphere, while SSA does not. 
 
\subsection{Clustering Properties}
\label{energy-function}

SSA normalises twice, over slots in the assignment weights and over inputs in the weighted mean.
Following \citet{geshkovski_mathematical_2025}, we remove the second to isolate the role of the first and compare to reveal the role of the second. 
Removing the weighted mean gives
\begin{equation}
    \label{eq:ussa}
    s^{t+1}_i = \sum_{j=1}^N w_{ij}^t x_j, \qquad w_{ij}^t = \frac{e^{\beta\langle s_i^t, x_j\rangle}}{\sum_{m=1}^M e^{\beta\langle s_m^t, x_j\rangle}},
\end{equation}
which we call unnormalised simplified slot attention (USSA). The energy function is given by 
\begin{equation}
    \label{eq:ussa_energy}
    E_1 = \frac{1}{2}\sum_{i=1}^{M}\|s_i\|^2
    -
    \sum_{i=1}^{M}\sum_{j=1}^{N}
        w_{ij}\langle s_i,x_j\rangle
    +
    \frac{1}{\beta}
    \sum_{i=1}^{M}\sum_{j=1}^{N}
        w_{ij}\log w_{ij},
\end{equation}
subject to \(\sum_i w_{ij}=1\) with \(w_{ij} > 0\) where alternating minimisation over \(w_{ij}\) and \(s_i\) gives USSA (see \ref{ussaderivation}).
Considering \(\beta\rightarrow\infty\), and minimising over the slot positions, the resulting energy is
\begin{equation}
    \label{eq:ussa-hard}
    E_1^*
    =
    -\frac12
    \sum_{i=1}^{M}
    \left\|
        \sum_{j=1}^{N}w_{ij}x_j
    \right\|^2,
    \qquad
    \sum_{i=1}^{M}w_{ij}=1,
\end{equation}
whose minima are attained at hard slot assignments where \(w_{ij} \in \{0, 1\}\), shown in \Cref{fixed-point}.
There we also show that splitting a cluster lowers this energy when the aggregate cross-cluster dot product is negative.
Thus, USSA's hard-assignment clustering geometry is governed by dot-product similarity rather than Euclidean distance.
For fixed slot positions, SSA and USSA induce identical assignments (\cref{eq:ssa,eq:ussa}), thus clustering points by the same criterion. However, USSA's weighted sum lets slots shrink to zero and die, while SSA's weighted mean prevents this.

\section{Experiments}
\label{experiments}

We now evaluate whether the competitive clustering dynamics isolated by SSA are sufficient for real-world object-centric segmentation.
\paragraph{Setup.}
We replace the slot attention segmentation module in the encoder-decoder architecture, outlined in \Cref{sec:slot_attn}, with $k$-means, SKM, SSKM, USSA and SSA.
Following \citep{seitzer_bridging_2023}, we evaluate these models on Pascal VOC 2012 \citep{everingham_pascal_2015} using frozen DINO input features \citep{caron_emerging_2021}.
Each method groups the same \(N=196\) patch input features (\(d=768\)) to \(M=6\) slots. 
We report FG-ARI and instance/class-level mean best overlap (\(\text{mBO}_i / \text{mBO}_c\)) from both the segmentation module's assignments and the decoder's alpha masks, and with patch reconstruction MSE.
Details are in \Cref{exp}.

\begin{table}[t]
    \centering
    \caption{
        Segmentation and reconstruction on Pascal VOC over 3 seeds.
        FG-ARI/mBO are computed from the segmentation assignments and decoder alpha masks.
        MSE measures DINO patch reconstruction.
        Best results are in \textbf{bold} and second-best results are \underline{underlined}.
    }
    \resizebox{\textwidth}{!}{%
    \begin{tabular}{l ccc ccc c}
        \toprule
        & \multicolumn{3}{c}{Clustering Assignment} 
        & \multicolumn{3}{c}{Decoder Assignment} 
        & Recon. \\
        \cmidrule(lr){2-4} 
        \cmidrule(lr){5-7} 
        \cmidrule(lr){8-8}
        Method 
        & FG-ARI $\uparrow$
        & mBO\textsubscript{i} $\uparrow$
        & mBO\textsubscript{c} $\uparrow$
        & FG-ARI $\uparrow$
        & mBO\textsubscript{i} $\uparrow$
        & mBO\textsubscript{c} $\uparrow$
        & MSE $\downarrow$ \\
        \midrule
        Slot Attn. (\cref{slot-attn-update})
        & $0.186 \pm 0.005$
        & $\mathbf{38.257 \pm 0.243}$
        & $\underline{40.790 \pm 0.226}$
        & $0.181 \pm 0.007$
        & $\mathbf{35.262 \pm 0.486}$
        & $\mathbf{37.232 \pm 0.469}$
        & $\mathbf{0.382 \pm 0.001}$ \\

        $k$-Means
        & $\underline{0.259 \pm 0.014}$
        & $30.533 \pm 0.518$
        & $35.558 \pm 0.573$
        & $\mathbf{0.225 \pm 0.012}$
        & $27.462 \pm 0.291$
        & $31.767 \pm 0.350$
        & $0.425 \pm 0.001$ \\
        SKM (\cref{eq:SKM})
        & $0.222 \pm 0.006$
        & $33.464 \pm 0.098$
        & $38.617 \pm 0.263$
        & $\underline{0.189 \pm 0.005}$
        & $29.869 \pm 0.281$
        & $34.075 \pm 0.158$
        & $0.417 \pm 0.001$ \\
        SSKM (\cref{SSKM})
        & $0.204 \pm 0.013$
        & $19.111 \pm 0.027$
        & $22.793 \pm 0.029$
        & $0.029 \pm 0.005$
        & $15.640 \pm 0.273$
        & $18.204 \pm 0.372$
        & $0.437 \pm 0.001$ \\
        USSA (\cref{eq:ussa})
        & $\mathbf{0.499 \pm 0.000}$
        & $19.159 \pm 0.003$
        & $22.990 \pm 0.003$
        & $0.095 \pm 0.010$
        & $22.166 \pm 0.297$
        & $25.635 \pm 0.277$
        & $0.464 \pm 0.000$ \\
        \midrule
        SSA (\cref{eq:ssa})
        & $0.155 \pm 0.001$
        & $\underline{37.369 \pm 0.379}$
        & $\mathbf{40.920 \pm 0.554}$
        & $0.147 \pm 0.005$
        & $\underline{32.214 \pm 0.339}$
        & $\underline{34.672 \pm 0.356}$
        & $\underline{0.390 \pm 0.001}$ \\
        \bottomrule
    \end{tabular}
    }
    \label{tab:results_1}
\end{table}

\begin{figure}[th]
    \centering
    \includegraphics[width=1.0\linewidth]{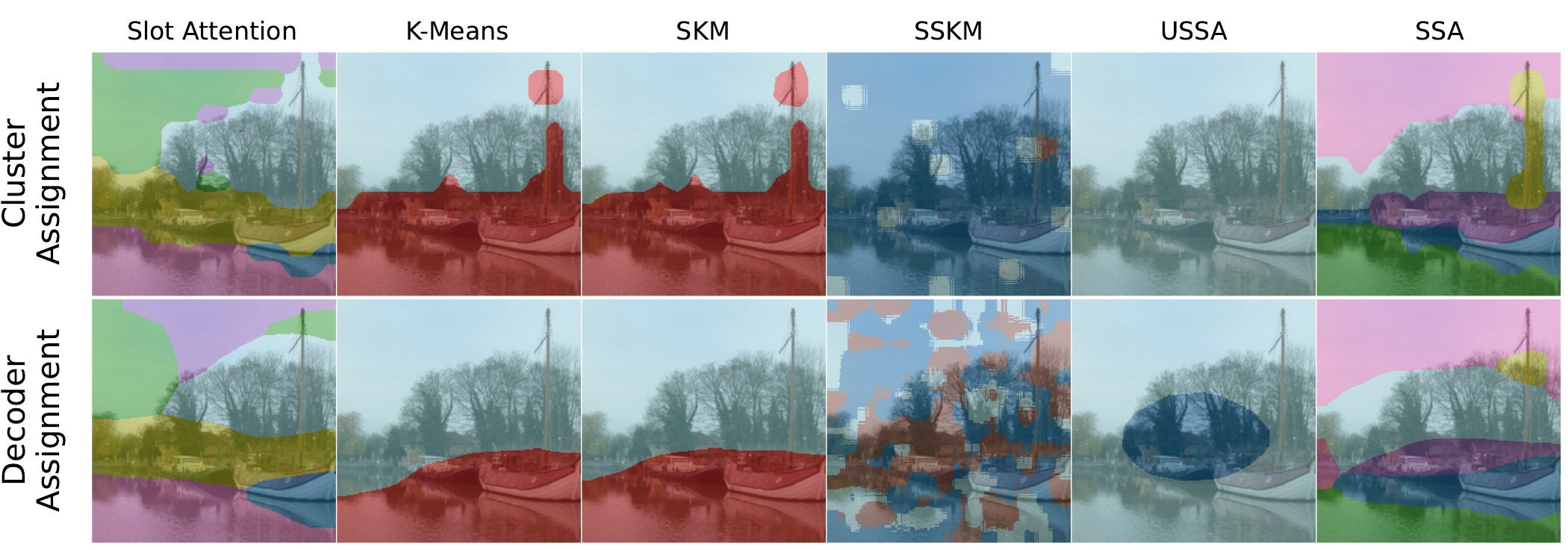}
    \caption{
    Pascal VOC segmentations across methods.
    Top: cluster assignments; Bottom: decoder alpha masks.
    Colours denote predicted slots and are not matched across methods.
    We provide additional qualitative examples in \Cref{app:more_results}.
    }
    \label{fig:exp1base}
\end{figure}

\paragraph{Results.}
The results in \Cref{tab:results_1,fig:exp1base} reflect the distinctions between the clustering dynamics identified in \Cref{relationship-to-kmeans,energy-function}.
SKM retains SSA's competitive weighted-mean update but uses Euclidean similarity, and tends to produce under-segmented groups.
SSKM instead shares SSA's assignment, but projects each updated centroid onto the unit sphere, which produces much poorer segmentations.
USSA retains SSA's instantaneous assignments but replaces the weighted mean with a weighted sum, and progressively collapses towards a small number of active slots (\Cref{app:ussa}).
Together, these comparisons suggest that both the dot-product assignments and weighted-mean update contribute to SSA's clustering behaviour.

SSA is best or second-best on every metric except FG-ARI, closely tracking slot attention.
FG-ARI should be interpreted with caution as it ignores background; the collapsed USSA (\Cref{fig:exp1base}) obtains the highest cluster-based FG-ARI.
Slot attention's slightly lower MSE suggests that learned components improve slot representations while simple dynamics suffice for clustering.

\section{Discussion}
\label{conclusion}
SSA combines the competitive weighted-mean update of soft $k$-means (SKM) with dot-product assignments, and both components matter because
SKM's Euclidean assignments produce coarser groupings, SSKM's spherical projection degrades segmentation, and USSA's weighted sum collapses the active slots.
On Pascal VOC with frozen DINO features, SSA performs close to slot attention with no learned components in the grouping module, suggesting slot attention's learned transformations operate on a clustering mechanism already present in its competitive attention structure rather than constructing it.

Our analysis has several limitations. Most importantly, the energy function
characterises the surrogate USSA dynamics rather than the weighted-mean
dynamics of SSA itself. Moreover, the empirical study is restricted to a single
dataset, frozen feature backbone, and reconstruction-based training setup, and
the roles of the individual components omitted from slot attention are only
discussed heuristically (\Cref{excluded-components}). Future work should
characterise the fixed points and convergence properties of SSA directly and
systematically reintroduce the omitted components to determine their respective
roles in damping, similarity learning, representation refinement, and
normalisation \citep{krimmel_attention_2024}. Evaluating these dynamics across datasets, feature
representations, and more demanding object-centric tasks would further clarify
the extent to which the observed behaviour generalises.

\bibliographystyle{plainnat}
\bibliography{references}

\newpage

\appendix

\section{Technical appendices and supplementary material}

\subsection{Additional Qualitative Results} \label{app:more_results}

\begin{figure}[h]
    \centering
    \subfigure[]{\includegraphics[width=0.95\textwidth]{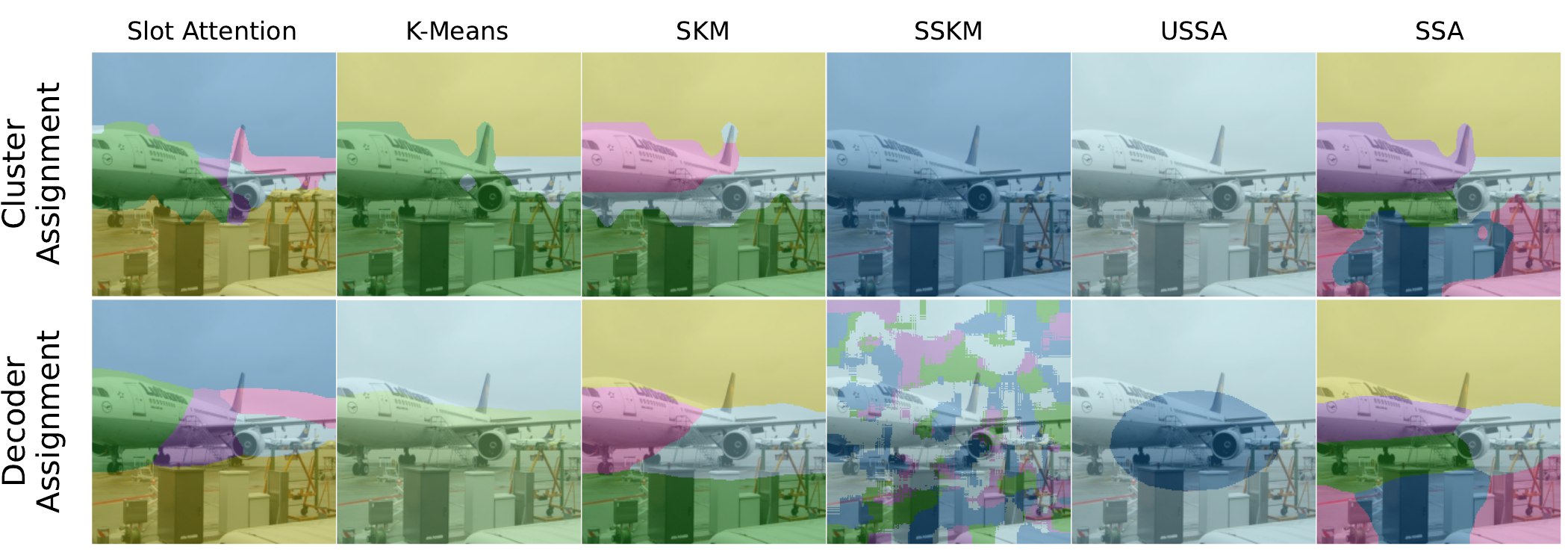}}
    \\
    \subfigure[]{\includegraphics[width=0.95\textwidth]{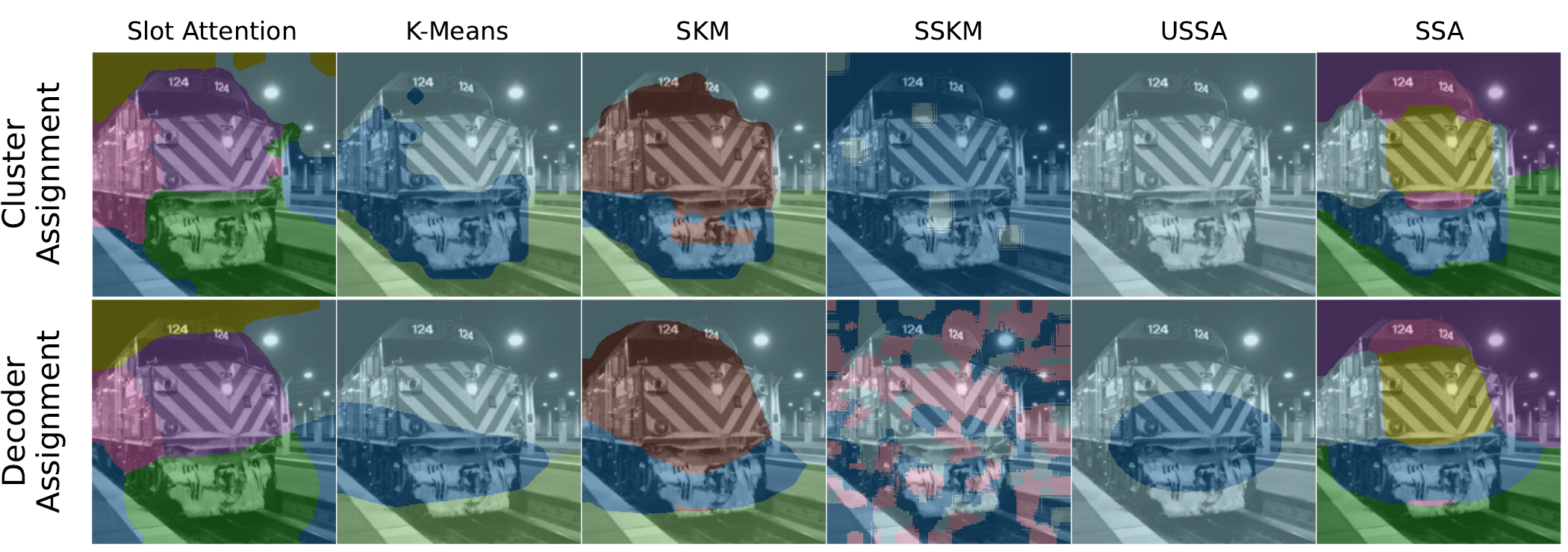}}
    \\
    \subfigure[]{\includegraphics[width=0.95\textwidth]{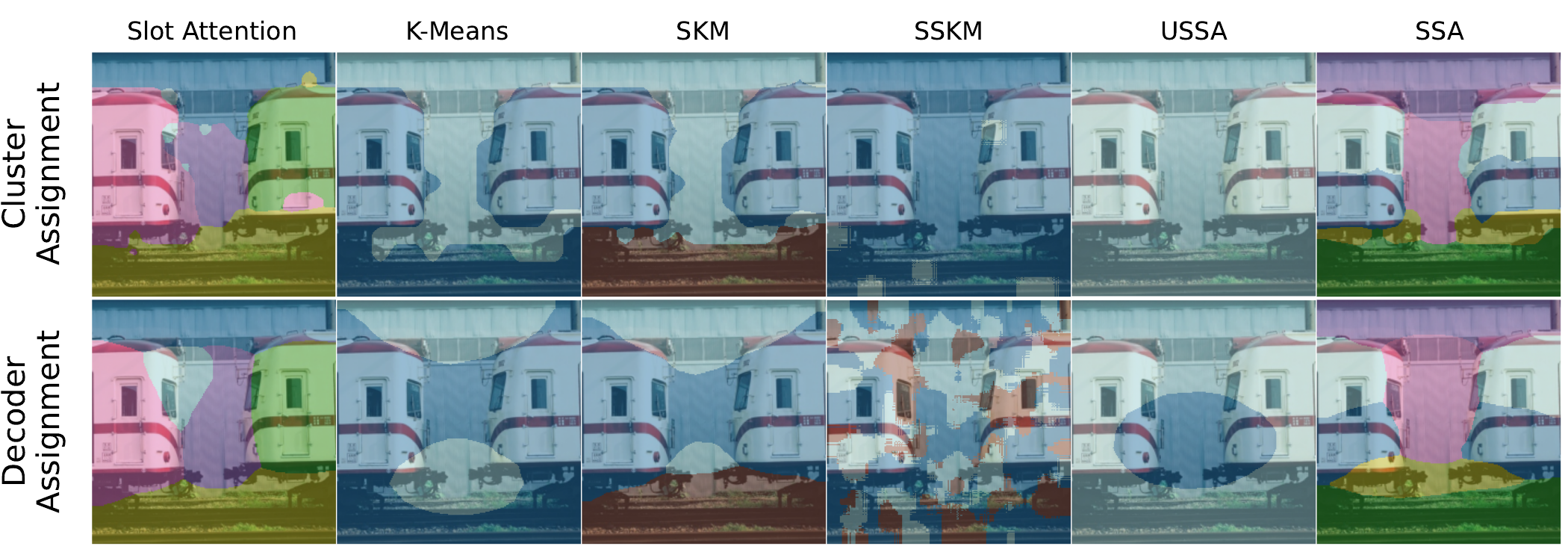}}
    \caption{Additional qualitative segmentation examples across all grouping modules, in the same format as \cref{fig:exp1base}.}
    \label{fig:app_more_examples}
\end{figure}

\subsection{Derivation of USSA Update} \label{ussaderivation}

The direct dot-product analogue of the soft $k$-means objective is
\begin{equation}
    \label{eq:unbounded-energy}
    E_n
    =
    -\sum_{i,j} w_{ij}\langle s_i,x_j\rangle
    +
    \frac{1}{\beta}\sum_{i,j}w_{ij}\log w_{ij}.
\end{equation}
Without a constraint on the slot norms, \cref{eq:unbounded-energy} is unbounded below. We therefore add an \(L^2\) regularisation term, giving
\begin{align}
    E_1 
    &= \frac{1}{2}\sum_{i=1}^M \|s_i\|^2 + \sum_{i=1}^M\sum_{j=1}^N -w_{ij}\langle s_i, x_j\rangle + \frac{1}{\beta} \sum_{i=1}^M\sum_{j=1}^N w_{ij}\log w_{ij} 
\end{align}

To find the fixed point of \(E_1\), we minimise with respect to the weights \(w_{ij}\) and the slots \(s_i\) separately, following the same two-step procedure used for soft $k$-means. We first form the Lagrangian associated with the column-normalisation constraint \(\sum_i w_{ij} = 1\),
\begin{equation}
\mathcal{L} = \sum_{i,j}\left[-w_{ij}\langle s_i, x_j\rangle + \frac{1}{\beta}w_{ij}\log w_{ij}\right] - \sum_j \lambda_j\left(\sum_i w_{ij} - 1\right).
\end{equation}
Differentiating with respect to \(w_{ij}\) and setting the result to zero gives
\begin{equation}
\frac{\partial \mathcal{L}}{\partial w_{ij}} = -\langle s_i, x_j\rangle + \frac{1}{\beta}\left(1 + \log w_{ij}\right) - \lambda_j = 0,
\end{equation}
which, after enforcing the normalisation constraint, recovers the familiar softmax weight update
\begin{equation}
w_{ij} = \frac{e^{\beta\langle s_i, x_j\rangle}}{\sum_{k=1}^K e^{\beta\langle s_k, x_j\rangle}}.
\end{equation}
Note that this is the same weight update obtained for \cref{eq:ssa}, since the quadratic regularisation term \(\frac{1}{2}\|s_i\|^2\) does not depend on \(w_{ij}\) and therefore vanishes under this differentiation.

Next, we fix the weights and minimise \(E_1\) with respect to \(s_i\). Differentiating gives
\begin{equation}
\frac{\partial E_1}{\partial s_i} = s_i - \sum_{j=1}^N w_{ij}x_j = 0,
\end{equation}
so that the optimal slot position is the weighted sum
\begin{equation}
s_i = \sum_{j=1}^N w_{ij}x_j.
\end{equation}
This is the USSA update introduced above.

\subsection{Fixed point of USSA}\label{fixed-point}

To find the fixed point, we can impose the equality \(s_i = \sum_{j}w_{ij}x_j\) into the energy to get
\begin{subequations}
    \begin{align}
    E^*_{1} &= \frac{1}{2}\sum_{i=1}^M \|s_i\|^2 + \sum_{i=1}^M\sum_{j=1}^N -w_{ij}\langle s_i, x_j\rangle + \frac{1}{\beta} \sum_{i=1}^M\sum_{j=1}^N w_{ij}\log w_{ij}\\
    &= \sum_{i=1}^M \frac{1}{2}\|\sum_{j=1}^N w_{ij}x_j\|^2 + \sum_{i=1}^M\sum_{j=1}^N -w_{ij}\langle \sum_{k=1}^N w_{ik}x_k, x_j\rangle + \frac{1}{\beta} \sum_{i=1}^M\sum_{j=1}^N w_{ij}\log w_{ij}\\
    &= \sum_{i=1}^M \left(\frac{1}{2}\|\sum_{j=1}^N w_{ij}x_j\|^2 - \langle \sum_{j=1}^N w_{ij}x_j, \sum_{j=1}^N w_{ij}x_j\rangle + \frac{1}{\beta} \sum_{j=1}^N w_{ij}\log w_{ij}\right)\\
    &= \sum_{i=1}^M \left(\frac{1}{2}\|\sum_{j=1}^N w_{ij}x_j\|^2 - \| \sum_{j=1}^N w_{ij}x_j\|^2 + \frac{1}{\beta} \sum_{j=1}^N w_{ij}\log w_{ij}\right)\\
    &= \sum_{i=1}^M \left(-\frac{1}{2}\|\sum_{j=1}^N w_{ij}x_j\|^2 +\frac{1}{\beta} \sum_{j=1}^N w_{ij}\log w_{ij}\right)
\end{align}
\end{subequations}
and in the zero temperature limit, ie. \(\beta \to \infty\), we have
\begin{align}
    E^*_{1} = -\frac{1}{2}\sum_{i=1}^M \|\sum_{j=1}^N w_{ij}x_j\|^2 \quad \text{subject to} \quad \sum_{i=1}^M w_{ij} = 1 \quad \text{for all j}
\end{align}

Since \(E_1^*\) is concave in the weights \(w_{ij}\) subject to \(\sum_i w_{ij}=1\), its minimisers lie on the vertices of the assignment simplex, so the optimal weights are either 0 or 1. Dropping the constant coefficient, which does not affect the location of the fixed points, gives
\begin{align}
    E^*_{1} = -\sum_{i=1}^M \left\|\sum_{j: w_{ij} = 1}x_j\right\|^2.
\end{align}
To understand the structure of these clusters, consider a partition of \(X\) such that \(X_1 \cup X_2 = X\) and \(X_1 \cap X_2 = \emptyset\), that is, splitting the data into two parts. We can then write
\begin{align}
    \|\sum_j x_j\|^2 =\|\sum_{x_j\in X_1} x_j\|^2 + \|\sum_{x_k \in X_2} x_k\|^2 + 2\sum_{x_j\in X_1, x_k \in X_2}\langle x_j, x_k\rangle.
\end{align}
This means a cluster is better split when the sum of its cross-cluster dot products is predominantly negative.

\subsection{USSA Evolution}
\label{app:ussa}

\begin{figure}[h]
    \centering
    \includegraphics[width=\linewidth]{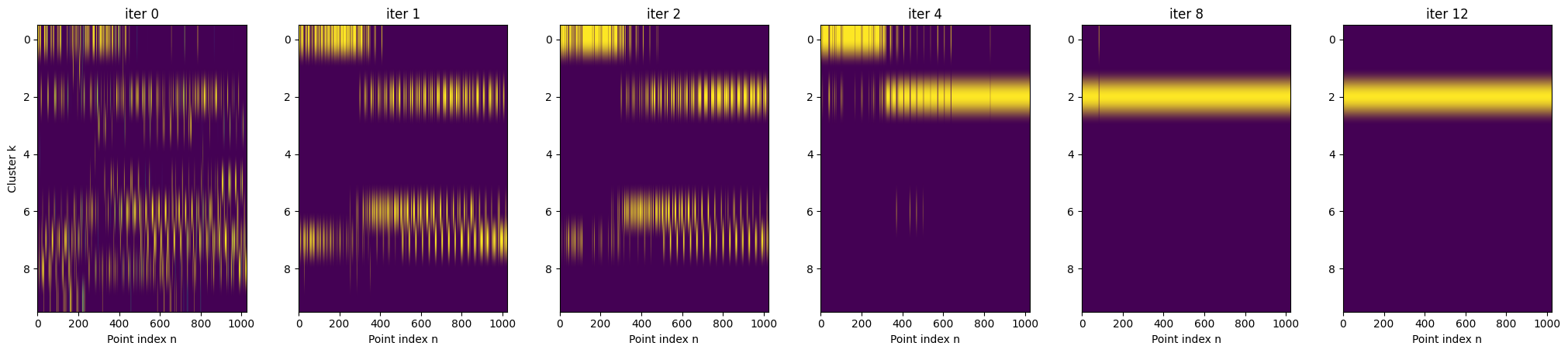}
    \caption{Evolution of the attention weight matrix $W \in \mathbb{R}^{K \times N}$ under the \cref{eq:ussa} update for $K=10$ slots and $N=1000$ gaussian initialised points across iterations. Bright entries indicate high weight $w_{ij}$. Slots progressively die out as the dominant slot accumulates mass, leaving a single active slot by iteration 12.}
    \label{fig:weight_evolution}
\end{figure}

\begin{figure}[h]
    \centering
    \includegraphics[width=\linewidth]{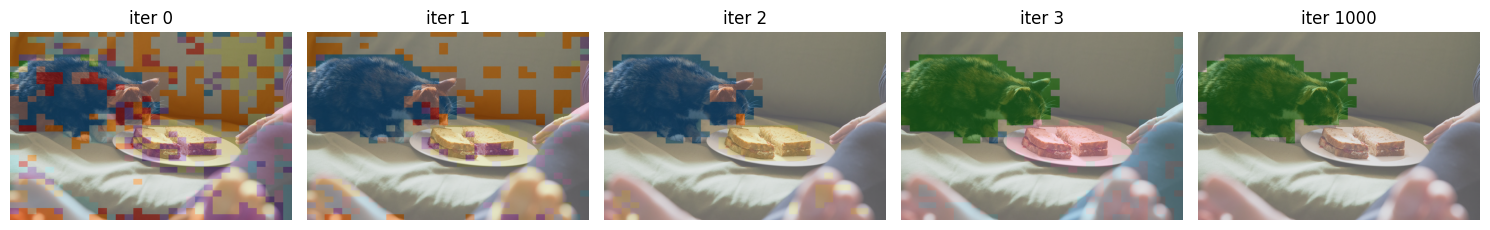}
    \caption{Evolution of segmentations under \cref{eq:ussa} where the segmentations are given by the cluster assignment. DINO features were extracted and \cref{eq:ussa} was run on these features. We see that by the end of the algorithm, only 2 slots survive, the cat and the background.}
    \label{fig:e1evol}
\end{figure}

\subsection{GRU, QKV and LayerNorm}\label{excluded-components}

In \cref{sec3}, we have provided and analysed a simplification of the slot attention algorithm. We have also shown that it performs clustering based on angle between slot and point and also magnitude of slot. In this section, we aim to provide some insight on the role of the previously excluded components. 

\paragraph{GRU} The gated recurrent unit is a parameterised function introduced by\citet{cho_learning_2014}. It was included by \citet{locatello_object-centric_2020} in the original slot attention as it yielded improved empirical results. The GRU update mixes the previous slot with a proposed update via a learned, dimension-wise interpolation, $s_i^{t+1} = (1-z_i^{t+1})\odot s_i^t + z_i^{t+1}\odot\hat{s}_i^{t+1}$, where the gate $z_i^{t+1}$ is itself a learned function of both. We hypothesise that this damping plays a role analogous to a gradient-descent step size. Taking a gradient step on $E_1$ with step size $\eta$ gives $s_i^{t+1} = (1-\eta)s_i^t + \eta\sum_j w_{ij}^tx_j$, which has the same damped-interpolation form as the GRU update, with $\alpha = 1-\eta$ matching a full step toward USSA's target update rather than SSA's. This suggests the GRU may be implementing something closer to a learned, partial gradient step on a bounded energy than the coordinate descent that produces SSA directly. This is a substantial simplification of the actual GRU, which additionally includes a reset gate and nonlinear activations not captured by this comparison, and the hypothesis is not tested empirically here.

\paragraph{Q, K, V} Beyond the identity case, the learned projections $Q,K$ can be heuristically understood as a learnt similarity function between slots and data points which is stronger than the simple dot product used in our simplification. In cases where it is desired to cluster points which are dissimilar according to dot product, it is useful to provide a learnt transformation which might more closely align two dissimilar points and hence group them together. The role of the V matrix is hypothesised to be closely linked to the downstream task where it is the transformation responsible for representing the slots. 

\paragraph{LayerNorm} Layer normalisation was introduced by \citet{ba_layer_2016} and normalises each slot to have zero mean and unit variance across its feature dimensions, followed by a learned elementwise rescaling and shift. In the context of our simplification, we hypothesise its role is closely related to the spherical \(k\)-means discussed in \cref{relationship-to-kmeans} and the sphere constrained dynamics of \citet{geshkovski_mathematical_2025}. It constrains slot growth without fully restricting slots to a unit sphere. As with the GRU, we do not test this hypothesis empirically here.

\subsection{Limitations}\label{limitations}

This analysis of slot attention as an interacting particle system has several limitations.

\paragraph{Incomplete simplified model} Slot attention was studied primarily through a single simplified, parameter-free variant (SSA). The precise role of the learned components, the GRU and the \(Q,K,V\) projections, remains only partially characterised from a mathematical perspective.

\paragraph{Small experiments} Experiments were conducted on a single dataset (Pascal VOC 2012) using a single frozen encoder (DINO ViT-B/16), so it is unclear how the conclusions generalise to other feature extractors or domains, and results were averaged over only 3 seeds, so the impact of randomness is not quantified. Reconstruction of DINO features is also a comparatively weak downstream task, so whether parameter-free clustering remains competitive on more demanding tasks such as generation or structured prediction remains open. 

\paragraph{USSA to SSA} The bounded energy \(E_1\) does not exactly correspond to SSA's own update, leaving a gap between the geometric intuition it provides and the algorithm actually evaluated. Additional work is required to characterise fixed points of \cref{eq:ssa} and identify convergence properties.

\subsection{Future Work}\label{future-work}

\paragraph{Energy based sampling} The energy functions derived in this work suggest a natural probabilistic reading of clustering. \citet{kulis_revisiting_2012} pursue a related idea from a Bayesian nonparametric perspective, reinterpreting \(k\)-means style updates as inference in a Dirichlet process mixture, which also allows the number of clusters to be inferred rather than fixed in advance. Extending this view to slot attention where we treat \(Q,K\) as parametrising a learned energy, and exploring sampling-based alternatives to gradient-based training such as contrastive divergence \citep{hinton_training_2002} is a promising direction for future work, and may also address the fixed-\(K\) assumption underlying both \(k\)-means and slot attention.

\paragraph{Further analysis of USSA} In additional experiments, the dynamics of \cref{eq:ussa} show interesting behaviour. As slots grow in magnitude due to the lack of self normalisation, we can empirically observe metastable clusters. As the algorithm progresses slots die out to 0 and each death takes a longer time. An example is shown in  \cref{fig:e1evol}. It would be interesting to check if the meta stable clusters in self attention correspond to the meta stable clusters under \cref{eq:ussa}.

\section{Implementation and Experiment Details}\label{exp}

\paragraph{Dataset and Preprocessing} All models were trained using PASCAL VOC 2012, with the \texttt{trainaug} split used for training and the \texttt{val} split used for evaluation and reporting. Images were resized to \(224\times224\) using bilinear interpolation; training additionally applied a random crop to \(224\times224\) and a random horizontal flip, while evaluation used a center crop. All images were normalised using ImageNet channel statistics. The frozen DINO ViT-B/16 encoder produces \(N=196\) patch tokens (a \(14\times14\) grid) of dimension \(d=768\), which are used directly as input to the grouping module, following \citet{seitzer_bridging_2023}, who first showed that grouping frozen DINO features, rather than training a feature extractor jointly with the grouping module, substantially improves slot attention's segmentation performance on real-world images.

\paragraph{Segmentation Modules} We compare six segmentation modules: $k$-means, soft $k$-means, SSA, and full slot attention. Each maps the \(N=196\) patch features \(X \subset \R^d\) (\(d=768\)) to \(M=6\) slot representations. \(k\)-means centroids, for both the regular and soft variants, are initialised randomly from \(\mathcal{N}(0,I)\) and updated for 20 iterations. Slot attention uses 3 routing iterations, with slots sampled each forward pass from a learned Gaussian initialised with zero mean and zero log-scale. SSA is implemented as a separate segmentation module that directly computes the update rule in \cref{eq:ssa}, with no \(Q,K,V\) projections, GRU, LayerNorm, or residual MLP, retaining only the weighted-mean aggregation step. All four segmentation modules share the same surrounding encoder and training pipeline, along with the decoder described below, so that any difference in performance can be attributed to the segmentation module itself, rather than to the training procedure.

\paragraph{Decoder Architecture} The decoder is a token-wise MLP: each slot is repeated across all \(N\) positions, added to a shared learned positional embedding, and passed through Linear(768\(\to\)768) \(\to\) LayerNorm \(\to\) Linear (768\(\to\)2048) \(\to\) ReLU \(\to\) [Linear(2048\(\to\)2048) \(\to\) ReLU]\(\times\)2 \(\to\) Linear (2048\(\to\)769), producing a 768-dimensional reconstruction and a scalar mask logit per slot per position. Slot and embedding dimensions were set equal (\(d_s = d = 768\)).

\paragraph{Training} The decoder was trained with AdamW (learning rate \(\eta_0 = 2\times10^{-4}\), weight decay \(10^{-6}\)), batch size 32, gradient-norm clipping at 1.0, for 150 epochs with evaluation every 5 epochs. The learning rate followed a linear warmup over the first 10{,}000 optimiser steps, followed by exponential decay with a half-life of 100{,}000 steps,
\begin{equation}
\eta(s) = \eta_0 \cdot \begin{cases} s / 10{,}000, & s < 10{,}000 \\ \left(0.5^{1/100{,}000}\right)^{s - 10{,}000}, & s \geq 10{,}000 \end{cases}
\end{equation}
This training recipe was identical across all segmentation-module variants; only the segmentation module itself differed between runs. All results are averaged over 3 seeds.

\paragraph{Evaluation and Metrics} We evaluate the same task and metrics as \citet{locatello_object-centric_2020}: the foreground-adjusted Rand index (FG-ARI), a metric measuring cluster similarity common in the object-centric literature, following \citet{seitzer_bridging_2023} in additionally reporting mean best overlap (mBO) \citep{pont-tuset_multiscale_2017} for real-world images. mBO is computed by assigning each ground-truth mask the predicted mask with the largest overlap, then averaging the IoU of the assigned pairs. Unlike FG-ARI, which only considers foreground pixels, mBO also accounts for background, and also measures how closely predicted masks fit object boundaries. As PascalVOC provides both instance-level and class-level masks, we report both instance-level mBO (mBO\(_i\)) and class-level mBO (mBO\(_c\)), which together reveal whether a method prefers instance- or semantic-level segmentations. Segmentation masks are extracted from either the segmentation module's own assignment weights or the decoder's alpha masks, upscaled from patch to pixel resolution via bilinear interpolation. We additionally report reconstruction MSE.

\paragraph{Compute} All models were trained on a single NVIDIA Tesla T4 GPU (15{,}360 MiB).

\end{document}